\documentclass[a4paper,conference]{IEEEtran}
\usepackage[utf8]{inputenc}
\usepackage{amsfonts}
\usepackage{amsmath}
\usepackage[pdftex]{graphicx}
\usepackage{url} 
\usepackage{hyperref}
\usepackage{cite} 
\usepackage{algorithmic}
\usepackage{array}
\usepackage{multirow}
\usepackage{balance}

\graphicspath{ {./figures/} }
\newcommand{\etal}{\textit{et al. }}

\begin{document}
\title{Dyadic Movement Synchrony Estimation Under Privacy-preserving Conditions}
\author{\IEEEauthorblockN{Jicheng Li}
\IEEEauthorblockA{Computer \& Information Sciences\\
University of Delaware\\
Newark, Delaware 19716\\
Email: lijichen@udel.edu}
\and
\IEEEauthorblockN{Anjana Bhat}
\IEEEauthorblockA{
Physical Therapy Department\\
University of Delaware\\
Newark, Delaware 19716\\
Email: abhat@udel.edu}
\and
\IEEEauthorblockN{Roghayeh Barmaki}
\IEEEauthorblockA{
Computer \& Information Sciences\\
University of Delaware\\
Newark, Delaware 19716\\
Email: rlb@udel.edu}}

\IEEEoverridecommandlockouts
\IEEEpubid{\makebox[\columnwidth]{***-*-****-****-*/**/~\copyright 2022 IEEE \hfill}
\hspace{\columnsep}\makebox[\columnwidth]{ }}
\maketitle
\IEEEpubidadjcol
\begin{abstract}
Movement synchrony refers to the dynamic temporal connection between the motions of interacting people.
The applications of movement synchrony are wide and broad. 
For example, as a measure of coordination between teammates, synchrony scores are often reported in sports.
The autism community also identifies movement synchrony as a key indicator of children's social and developmental achievements.
In general, raw video recordings are often used for movement synchrony estimation, with the drawback that they may reveal people's identities.
Furthermore, such privacy concern also hinders data sharing, one major roadblock to a fair comparison between different approaches in autism research.
To address the issue, this paper proposes an ensemble method for movement synchrony estimation, one of the first deep-learning-based methods for automatic movement synchrony assessment under privacy-preserving conditions.
Our method relies entirely on publicly shareable, identity-agnostic secondary data, such as skeleton data and optical flow. 
We validate our method on two datasets: 
(1) PT13 dataset collected from autism therapy interventions and (2) TASD-2 dataset collected from synchronized diving competitions.
In this context, our method outperforms its counterpart approaches, both deep neural networks and alternatives. 
\end{abstract}


\ifCLASSOPTIONpeerreview
\begin{center} \bfseries EDICS Category: 3-BBND \end{center}
\fi
%
\IEEEpeerreviewmaketitle

\section{Introduction}
In social psychology, \emph{movement synchrony (MS)} refers to the degree to which individuals move in similar ways over time~\cite{paxton2017interpersonal}. 
The applications of movement synchrony estimation are wide and broad.
For example, judges rate the synchrony between two divers in synchronized diving competitions.
In autism treatment, movement synchrony is a significant criterion because it reflects coherence between therapists and children with autism, revealing their physical and physiological development achievements.
A method to estimate movement synchrony automatically has received widespread attention \cite{ramseyer2020motion,georgescu2020reduced,altmann2021movement,Li2021improving,gao2020asymmetric}. 
However, privacy issues were not given a high priority in previous works.
Video recordings carrying critical visual and acoustic identifiers cannot be publicly accessible in areas where data privacy is crucial, such as patient behavior analysis.
The shortage of benchmark datasets restricts the application of cutting-edge machine learning techniques in turn. 

\begin{figure*}
    \centering
    \includegraphics[scale=0.05]{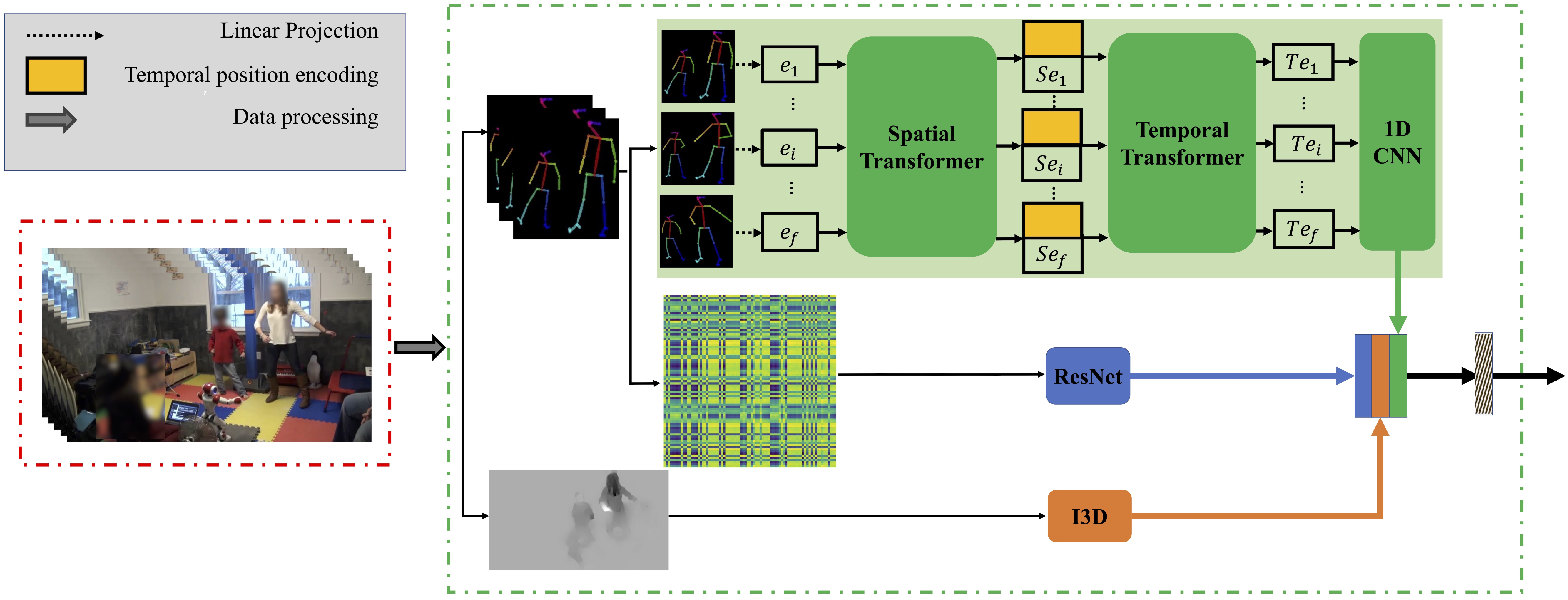}
    \caption{The proposed framework for movement synchrony estimation from video. 
    $e_{i} $, $Se_{i}$, and $Te_{i}$ represent the embedding generated from a linear projection layer, a spatial transformer, and a temporal transformer, respectively, while the input is the joint coordinates acquired from video by a human pose detector. Spatial position encoding in the spatial transformer is omitted. Dashed red (green) box: privacy-violating (privacy-preserving) components.
    Data processing (solid grey arrow) is handled by certificated experts and professionals to ensure that original data is hidden from the framework and only anonymized, privacy-preserving secondary data is granted access. As long as the original video data is not accessible by the framework, privacy is guaranteed.
    }
\vspace{-10pt}    
    
    \label{fig:framework}
\end{figure*}

Therefore, to tackle this challenge, we proposed a new problem: \textbf{to automatically estimate movement synchrony under privacy-preserving conditions}.
We also provided a privacy-preserving solution, one of the first deep-learning-based methods for determining movement synchrony in this regard.
Our approach depends entirely on identity-agnostic and privacy-preserving secondary data while maintaining crucial body movement features for motion understanding. 
The proposed framework (see Fig. \ref{fig:framework}) is an ensemble network consisting of three major components: 
(1) a skeleton-based spatial-temporal transformer network, 
(2) an optical flow based 3D convolutional networks \cite{carreira2017i3d}, and (3) a 2D convolutional neural network accepting temporal similarity matrix as inputs. 

Transformer Network (TFN) \cite{vaswani2017attention} is a novel neural network architecture based on the self-attention mechanism to draw global dependencies of sequential data. 
Inspired by \cite{zheng2021poseformer,plizzari2021skeletonSTTF}, in this work, we applied a spatial transformer to encode the local relationships between body joints and a temporal transformer to capture global dependencies across frames.
An optical flow is commonly referred to as the apparent motion of individual pixels between two consecutive frames on the image plane. 
Optical flow derived from raw videos can provide a concise description of both the region and velocity of a motion without exposing an individual's identity
\cite{carreira2017i3d,Feichtenhofer2016twostream,gao2020asymmetric,pandey2020guided}. 
A temporal similarity matrix (TSM) is a graphical representation to compare sequential data. 
The similarity can be determined by a pre-defined similarity function, such as Euclidean distance. 
TSMs have been used widely in human motion analysis tasks due to their robustness against perspective shifts and superior generalization abilities \cite{sun2015exploring,su2021how,dwibedi2020counting,panagiotakis2018unsupervised}. 
This paper computes the similarity between two skeleton sequences, either from one child and one therapist or two athletes participating in synchronized diving. 
Such a temporal similarity matrix is referred to as ``Cross-Similarity Matrix'' (CSM), in contrast to self-similarity matrix (SSM) \cite{sun2015exploring,su2021how,dwibedi2020counting,panagiotakis2018unsupervised}.

To justify our work, we performed experiments on two datasets: (1) PT13 dataset \cite{Li2021improving}, a real-world dataset on autism therapy, and (2) TASD-2  dataset \cite{gao2020asymmetric}
collected from synchronized diving competitions.
Our method surpassed its competitors for movement synchrony estimation without mandating access to the original videos, which contain sensitive personal information.
In summary, our contributions are:
(1) We introduced a new problem: to automatically estimate movement synchrony under privacy-preserving conditions;
(2) To solve the problem, we proposed a deep-learning-based framework, one of the first solutions for movement synchrony assessment under privacy-preserving conditions. Our framework is entirely based on identity-agnostic and privacy-preserving secondary data, such as skeleton data and optical flow.
Experiment results show that our approach outperforms its counterparts on both PT13 and TASD-2 datasets, including approaches with and without deep neural networks. 
    


\section{Related Work}
\label{sec:relatedwork}
\subsection{Movement Synchrony Estimation From Video}

Existing approaches can be mainly divided into two categories: (1) statistical approaches and (2) deep learning approaches.
Statistical approaches are based on low-level pixel-wise information, whereas deep learning approaches emphasize high-level semantics.

Representing statistical studies \cite{schoenherr2019quantification,ramseyer2020motion,georgescu2020reduced,altmann2020associations,altmann2021movement, schoenherr2019nonverbal} widely applied frame-differencing methods such as motion energy analysis (MEA) to quantify movement synchrony in a video.
Frame-differencing is accomplished by comparing successive static images, subtracting their pixel values, and summing the pixel changes between the two images.
However, MEAs restrict people's interaction to a pre-defined region of interest (ROI); thus they are noise sensitive and will fail upon scenes when individuals move out of the ROI.
Furthermore, some crucial semantic information is overlooked in the computation of frame differences, including the direction and form of movements \cite{ramseyer2020motion}.
Interestingly, MEA was compared with skeleton-based approach for nonverbal synchrony estimation in \cite{fujiwara2021video}. 
The study found that although two methods produced consistent results, the skeleton-based approach has an advantage in pinpointing specific body parts where synchrony exists.
Other works such as  \cite{chu2015unsupervised,chu2017branch, gashi2019using,Chang_2019_CVPR, sariyanidi2020discovering} investigated synchrony via pixel-level knowledge.
For example, some works \cite{gashi2019using,Chang_2019_CVPR} applied dynamic time warping (DTW) \cite{berndt1994dtw} for synchrony discovery and alignment.
In \cite{olsen2018simultaneous}, DTW was used to align multivariate functional curve data (for example, walking trajectories) and estimate their cross-covariance. Evangelos~\etal~\cite{sariyanidi2020discovering} used motion magnitude of pixels to find the largest groups that are moving together. 


Deep learning approaches can leverage semantic knowledge better than statistical approaches due to their superior capabilities in feature extraction and representation learning \cite{lecun2015deep}. 
Calabr{\`o} \etal \cite{calabro2021mms} reconstructed the Inter Beat Intervals (IBI) segments of electrocardiogram data via a convolutional auto-encoder to explore dyadic interaction between children and therapists. 
Li \etal \cite{Li2021improving} proposed a multi-task framework to integrate movement synchrony estimation with auxiliary tasks such as intervention activity recognition and individual action quality assessment. 
Nevertheless, both works \cite{calabro2021mms,Li2021improving} necessitate access to original video recordings, making them non-privacy-preserving. 

We also notice that many works investigated synchrony in biomedical signals such as electroencephalogram (EEG) using statistical methods \cite{wang2006phase, grinsted2004cwt,coco2014cross, parnamets2020physiological}.
In spit of that, these methods remain underexplored in the estimation of movement synchrony from videos. 
One possible reason is that video data (including the secondary data generated from the video) and biomedical signals are intrinsically different, especially when it comes to understanding human activities.

\subsection{Privacy-preserving Machine Learning }
Steil \etal \cite{steil2019privaceye} presented a way to automatically disable an eye tracker's first-person camera through a mechanical shutter when the user's privacy is jeopardized.
Singh \etal \cite{singh2021human} investigated human attributes (emotion, age, and gender) prediction under various de-identification privacy scenarios by body parts obfuscation.
Vu \etal \cite{vu2020privacy} designed a framework for social media content tagging by constructing a global knowledge graph to avoid sensitive local data such as faces, passport numbers, and vehicle plates.
However, these works primarily focused on static images, and none of them considered privacy issues in human motion evaluation, such as movement synchrony.
Further, our approach emphasizes context and background privacy since events like play therapy interventions are commonly held at children's homes, and the input data (skeleton data and optical flow) our model utilizes can mitigate such concerns.

\subsection{Skeleton-based Transformer Networks}
Transformer Networks \cite{vaswani2017attention,devlin2019bert,zhou2021informer,Arnab2021videotf} are extremely powerful in manipulating long sequential data. 
Previous studies have applied skeleton-based TFNs to human activity understanding.
A skeleton-based transformer was proposed in Plizzari'work \cite{plizzari2021skeletonSTTF} for activity recognition. 
The network relies on one spatial self-attention module to capture intra-frame body parts relations and one temporal self-attention module to model inter-frame correlations.
Zheng \etal \cite{zheng2021poseformer} proposed a transformer-based approach for 3D pose estimation from 2D videos. 
The model predicts the 3D pose within each (center) frame given its neighboring 2D frames.
Inspired by Zheng's work \cite{zheng2021poseformer}, we adapted their spatial-temporal transformer as an encoder, and revised the embedding layer to accommodate dyadic skeletons instead of one. 

\subsection{Temporal Similarity Matrix}
TSMs have been widely used in human action comprehension \cite{sun2015exploring,su2021how,dwibedi2020counting,panagiotakis2018unsupervised}.
Panagiotakis \etal \cite{panagiotakis2018unsupervised} used a temporal similarity matrix to find all periodic segments of a video in an unsupervised manner.
Dwibedi \etal \cite{dwibedi2020counting} applied a self-similarity matrix, as one type of TSM, to discover periodic events in a sequence by evaluating similarities between components.
Based on SSM, Nam \etal \cite{nam2021zero} developed an event proposal module to cluster and propose event regions while integrating global contextualization information.
Su \etal \cite{su2021how} proposed ``RhythmicNet'', which takes a video of human actions as input and generates a soundtrack for it. 
They used an SSM to capture body dynamics and regularize the transformer encoder. 
While SSM measures similarity within one sequential data, this paper introduced the ``Cross-Similarity Matrix" to determine the degree of similarity between two sequences, for example skeletons from two individuals.

\section{Method}
\label{sec:method}
The proposed framework (Fig. \ref{fig:framework}) has three major components:
\begin{enumerate}
    \item TFN branch: skeleton-based spatial-temporal transformer network;
    \item I3D \cite{carreira2017i3d} branch: optical flow based 3D convolutional network (I3D);
    \item CSM branch: ResNet \cite{he2016resnet} that accepts temporal similarity matrix as inputs. 
\end{enumerate}

\subsection{Spatial-temporal Transformer}
We adapted a spatial-temporal transformer as the encoder for dyadic skeleton sequences inspired by of Zheng's work \cite{zheng2021poseformer}.
The spatial transformer module generates a hidden embedding for one static frame, whereas the temporal transformer module  captures global dependencies across frames. 
We define a \textit{pose} is composed of $J$ \textit{joints} for each person, therefore each \textit{dyadic pose} $p$ from two interacting people is composed of $J \cdot 2$ joints.
Given that each joint $j \in \mathbb{R}^{2} $, we have $p \in \mathbb{R}^{J \cdot 4}$. 
To begin with, we will explain the attention mechanism of transformer networks first, followed by insight into the structure of spatial and temporal modules. 

\textbf{Scaled Dot-Product Attention} is one of the most commonly used attention functions in practice due to its benefits from parallel computing.
We can denote an input as $X \in \mathbb{R}^{f \times d}$, where $f$ is the sequence length and $d$ represents hidden feature dimension. 
Instead of computing attention directly from $X$, we introduce three matrices generated from $X$ (\textit{Eq.} (\ref{eq:qkv})): a query matrix $Q$, a key matrix $K$ and a value matrix $V$. 
Intuitively, query $Q$ describes the data we are probing, key $K$ reflects the query's relevancy, and value $V$ represents the input's intrinsic content.
$Q$, $K$ and $V$ can be computed by three unique linear projection functions: 
\begin{equation}
    Q = XW_Q,\quad K=XW_K,\quad V=XW_V
    \label{eq:qkv}
\end{equation}
where $Q,K,V \in \mathbb{R}^{f \times d}$, and $W_Q, W_K, W_V \in \mathbb{R}^{d \times d} $.
Therefore, the dot-product attention is calculated using the following formula:
\begin{equation}
     \operatorname {Attention}(Q, K, V)=\operatorname{Softmax} \left(\frac{Q K^{\top}}{ \sqrt{d}}\right) V
    \label{eq: att}
\end{equation}
An empirical \textit{scaling factor} $\sqrt{d}$ is introduced in Eq. (\ref{eq: att}) to ensure a stable training process \cite{vaswani2017attention}.

\textbf{Multi-head Self-Attention (MHSA)}
is a self-attention mechanism that correlates different positions in a sequence to create a representation of the sequence \cite{vaswani2017attention}.
Instead of computing self-attention directly from $Q, K$ and $V$, the attention is calculated in parallel on a subset of $Q, K$ and $V$, and each separately computed attention is known as a "head" (see \textit{Eq.}(\ref{eq:one_head})). 
As the number of heads $h$ is usually greater than one (single-head attention otherwise), the derived attention is thus referred to as \textit{"multi-head" self-attention}.
Empirically, we set $h$ to $8$ \cite{zheng2021poseformer,vaswani2017attention,Giuliari2020TFTrajectory,plizzari2021skeletonSTTF}. 
Multi-head self-attention allows the model to simultaneously attend to information originating from different representation sub-spaces located at different locations, while single-head attention fails to accomplish this \cite{vaswani2017attention}.
All heads are concatenated together eventually to generate the final output (\textit{Eq.}(\ref{eq:multi_head_attn})), $W_{out} \in \mathbb{R}^{d \times d} $, $head_i \in \mathbb{R}^{f \times d_{head}}, d_{head} = d // h$.
\begin{align}
    \operatorname{MHSA}(Q, K, V) =\operatorname{Concat}\left(head_{1}, \ldots, head_{h}\right) W_{\text {out }} \label{eq:multi_head_attn} \\
  \quad head_{i} =\operatorname{Attention}\left(Q_{i}, K_{i}, V_{i}\right), i \in \{1, \ldots, h \}
    \label{eq:one_head}
\end{align}

\textbf{Spatial Transformer} generates a hidden spatial embedding from joints detected within each frame. 
The spatial embedding has two components: (1) a patch embedding $E$ and (2) a positional embedding $E_{pos}$.
The patch embedding represents the hidden features of the input (i.e., joint coordinates), while the positional embedding retains positional information of the sequence (i.e., the order of joints).
In this work, we generate patch embedding $E$ by mapping input $X$ to a high dimensional feature space $Z$ via a linear projection $LP(\cdot)$, and adopt a trainable positional embedding $E_{pos}$ \cite{devlin2019bert,zheng2021poseformer} rather than pre-defined positional encoding rules \cite{vaswani2017attention}.   
The positional embedding is eventually added to the patch embedding, which can be formulated as:
\begin{align}
    Z &= [E_1; E_2; \ldots ; E_f] + E_{pos} \\
    \text{where} \quad  E_i & = ST(z_i), z_i = LP(x_i),
    i \in \{1, \ldots, f \}
\end{align}
$ST(\cdot)$ represents one spatial transformer  \textit{layer}, $Z, E_{pos} \in \mathbb{R}^{f \times c_{spa}} $, $c_{spa}$ is the spatial feature dimension in hidden space.
Due to the fact that spatial transformer is composed of a stack of $L$ identical \textit{layers}, steps (5) and (6) will be repeated $L$ times to generate $Z_L$, where $Z_l^{i}$ represents the hidden vector of frame $i$ generated from the $l$-th layer, $l \in \{1, \ldots,L\}$. 

\textbf{Temporal Transformer} models frame dependencies across the sequence. 
The temporal transformer takes the final output $Z_L$ from the spatial transformer as input and generates a temporal patch embedding $ET$.
A trainable temporal positional embedding \cite{dosovitskiy2020vit} $ET_{pos} \in \mathbb{R}^{f \times c_{temp}}$ is added to $Z_L$ to retain relative position information across frames, $c_{temp}$ is the temporal feature dimension. 
Same as the spatial transformer, the temporal transformer is also composed of a stack of $L$ identical \textit{layers}, and its output is denoted as  $Y \in \mathbb{R} ^{f\times c_{temp}}$. Let $Temp(\cdot)$ represent one temporal transformer layer:
\begin{align}
   Y_{k+1}  &= [ET_{k+1}^1; ET_{k+1}^2; \ldots ; ET_{k+1}^f] + ET_{pos} \\
    ET_{k+1}^i  &= Temp(Y_k^i)
\end{align}
$k \in \{0, \ldots, L-1\}, i \in \{1, \ldots, f\}$. Specifically, $Y_0  = X_L$.


\subsection{3D Convolutional Neural Networks} 
A standard method of comprehending human activities in videos involves using 3D convolutional neural networks that perform convolutions across spatiotemporal video volume \cite{tran2015c3d,pan2019action,carreira2017i3d,Feichtenhofer2016twostream}. 
Due to privacy concerns, the proposed network is not allowed access to the original RGB videos. 
Instead, in this paper, we adopted a flow-steam Inflated 3D ConvNet (I3D) \cite{carreira2017i3d} pre-trained on Kinetics dataset \cite{kay2017kinetics}, which only takes optical flows as input.
I3D is based on 2D ConvNet inflation, where filters and pooling kernels are expanded into three dimensions, making it possible to learn spatio-temporal features from video while leveraging successful ImageNet \cite{deng2009imagenet} architecture designs and parameters.

\subsection{Cross Similarity Matrix}
A temporal similarity matrix can provide significant insights into capturing both the spatial and temporal dynamics of a sequence.
In this paper, we computed \textit{cross similarity matrices} from skeleton sequences of interacting dyads. 
Given two skeleton sequences $X_1, X_2 \in \mathbb{R}^{f \times (J \cdot 2)}$, where $f$ is the sequence length, $J$ is the number of joints.
$X_p^t \in  \mathbb{R}^{(J \cdot 2)} $ represents the pose at the $t$-th time stamp of person $p$, 
and $X_p^t[k] \in \mathbb{R}^2$ denotes the 2D coordinates of joint $k$, 
where $t\in \{1, 2, \ldots, f\}, p \in \{1,2\}$, $k \in \{1,2, \ldots, J\}$. 
Therefore, a CSM can be expressed by a $f \times f$ matrix, and $CSM[i][j]$ describes the similarity between pose $X_1^i$ from the first person at time stamp $i$, and pose $X_2^j$ from the second person at time stamp $j$, respectively, $i,j \in \{1, 2, \ldots, f\}$. 
All 2D coordinates are normalized to range $[0,1]$ based on the image size. Consequently, we have
\begin{equation} %
    CSM[i][j] =  - \frac{1}{J} \sqrt{\sum_{k=1}^{J} \| X_{1}^{i}[k] - X_{2}^{j}[k]} \|^{2}
    \label{eq:csm_raw}
\end{equation}

\section{Dataset}
\label{sec:dataset}
We evaluated our framework on two datasets: \textit{Play Therapy 13 dataset (PT13)} \cite{Li2021improving} and \textit{Two-person Action Synchronized Diving dataset (TASD-2)} \cite{gao2020asymmetric}. 

\textbf{PT13} is derived from video recordings of therapy interventions for children with autism \cite{bhat2020play, Li2021improving}. 
It covers 13 unique activities, such as jumping, drumming, and squatting.
PT13 consists of 1,273 data samples as indicted in Table \ref{tab:pt13}, and all samples are categorized into three classes based on the  level of synchrony between children and therapists: \textit{Synchronized} (Sync), \textit{Moderated Synchronized} (ModSync) and \textit{Unsynchronized} (Unsync) . 
While the Synchronized (Unsynchronized) reflect absolute (no) synchrony between children and therapists, Moderated Synchronized refers to children reacting appropriately despite certain defects such as slight inconsistency or tardiness.
The average data length is 148 frames, with dimensions ranging from 320 by 240 pixels to 720 by 480 pixels.

\begin{table}[!t]
\caption{Statistics of PT13 dataset.}
\label{tab:pt13}
\centering
\begin{tabular}{|c|c|c|c|}
\hline
PT13 & Train & Test & Total (n, \%) \\
\hline
Synchronized & 345 & 87 & 432 (33.94\%)\\
\hline
Moderated Synchronized & 352 & 89 & 441 (34.64\%) \\
\hline
Unsynchronized & 320 & 80 & 400 (30.42\%) \\
\hline
Total & 1017  & 256  & 1273\\
\hline
\end{tabular}
\end{table}
\textbf{TASD-2} is composed of video recordings collected from two synchronized diving events: \textit{synchronized 3-m springboard} and \textit{synchronized 10-m platform}.
All videos are normalized to 102 frames long with a resolution of $320 \times 240$. 
By discarding six videos where the human pose detector \cite{cao2019openpose} failed to work, we had a dataset of 600 samples, 238 of which were in synchronized 3-m platform and 362 in synchronized 10-m platform. 
We adopted the train/test split as \cite{gao2020asymmetric}. 
In contrast to \cite{gao2020asymmetric}, we rely solely on the ``synchrony score" of TASD-2 in this study as the primary objective is synchrony estimation.
The synchrony score ranges from 0 to 10, with a mean of 7.64 and a standard deviation of 0.91. 


\textbf{Data Processing} was handled by certificated therapists and professionals, and the original data is kept completely confidential, as illustrated in Fig. \ref{fig:framework}.
To comply with the privacy principles, the proposed framework cannot accept original video recordings as inputs.
Instead, we extracted and computed secondary data, including skeleton data, optical flow, and CSMs from the video.
By using a human pose detector, joint coordinates were obtained from ``valid" frames, and one frame may only be valid if both persons can be detected; otherwise, it is invalid (zero or one detection).
Invalid frames were discarded eventually.
All data samples were uniformly sampled and normalized to 81 frames following \cite{zheng2021poseformer}. 
Optical flows and CSMs were resized to $224 \times 224$ by the nearest neighbor interpolation strategy. 
Upon completion of the data processing stage, only the secondary data in accordance with privacy principles were granted access to the framework.

\section{Experiments}
\label{sec:experiment}

\subsection{Implementation Details}
We implemented our proposed framework with Pytorch \cite{paszke2019pytorch}. Four NVIDIA Tesla V100 GPUs were used for training and testing. 
For both spatial and temporal transformer, we set $L =4$, $c_{spa} = c_{temp} = 544$.
We chose a sequence length $f = 81$ as \cite{zheng2021poseformer} and set $J = 17$ following COCO \cite{lin2014coco} format.
We trained our model using the Adam optimizer \cite{Diederik2015adam} for 800 epochs and set the batch size to 64. 
We adopted an exponential learning rate decay scheduler with the initial learning rate of 1e-3 and a decay factor of 0.98 for each epoch.
The dropout \cite{srivastava2014dropout} rate was set to $0.5$ in training phase to prevent overfitting. 
We applied OpenPose \cite{cao2019openpose} for 2D human pose detection, and optical flows were computed by TV-L1 algorithm \cite{Sanchez2013tvl1}.

\subsection{Ablation Study and Performance Comparison}
To verify the contribution of each branch of our framework, we conducted extensive ablation experiments on both TASD-2 and PT13 datasets.
Table \ref{tab:ablation_pt13} reports the recall of synchrony classification on PT13, and Table \ref{tab:ablation_tasd2} presents the mean square error (MSE) on TASD-2 for synchrony score prediction.  ($\uparrow$) indicates that larger values are better for recall in 
Table \ref{tab:ablation_pt13}, and  ($\downarrow$) indicates that a smaller value is better for MSE in 
Table \ref{tab:ablation_tasd2}.
In terms of single model evaluation, TFN achieved the highest classification accuracy of $91.46\%$ on PT13, whereas I3D performed best on TASD-2 with an MSE of 0.405. 
However, I3D did not perform well on PT13 with an accuracy slightly above random guess. 
Therefore, we excluded I3D in the following experiments on PT13.
In general, an ensemble network outperforms each individual sub-network.

 In addition, we also compared our method with other approaches \cite{berndt1994dtw,Pedregosa2011scikit-learn,coco2014cross, gao2020asymmetric} with and without using deep neural networks. 
 In summary, our method outperforms its counterparts on both datasets.
 Specifically, \cite{berndt1994dtw,Pedregosa2011scikit-learn,coco2014cross} generated an output by only accepting skeleton data as input, and the final score/class prediction is made by a support vector machine \cite{cortes1995svm} given the generated output.
 To provide a fair comparison, TFN accepting skeleton data only already outperforms \cite{berndt1994dtw,Pedregosa2011scikit-learn,coco2014cross}.


\begin{table}[!t]
\caption{Ablation study on PT13 reported by accuracy (\%). A higher value $\uparrow$ is better.}
\label{tab:ablation_pt13}
\centering
\begin{tabular}{|c||c|c|c|c|}
\hline
 & Sync & ModSync & Unsync & Avg.\\
\hline
DTW \cite{berndt1994dtw} & 79.31& 76.40 & 87.50 &80.86\\
\hline
2D Correlation \cite{Pedregosa2011scikit-learn} & 48.28 & 37.08 & 22.50 & 36.33\\
\hline
Cross-recurrence \cite{coco2014cross} & 86.21 & 71.91 & 16.25 & 59.38\\
\hline
AIM \cite{gao2020asymmetric} & 62.58 & 35.66 & 40.17 & 46.14\\
\hline
\hline
TFN  &94.25 & \textbf{86.52} & \textbf{93.75} & 91.41 \\
\hline
I3D \cite{carreira2017i3d} & 64.37 & 29.21 & 31.25 & 41.80 \\
\hline
CSM \cite{he2016resnet}& \textbf{97.70} &78.65 &91.25 & 89.06\\
\hline
TFN + I3D & \multicolumn{4}{c|}{$\mbox{---}$} \\
\hline
TFN + CSM  &\textbf{97.70} &85.39 & 92.50 &\textbf{91.80} \\
\hline
I3D + CSM & \multicolumn{4}{c|}{$\mbox{---}$}  \\
\hline
TFN+I3D+CSM & \multicolumn{4}{c|}{$\mbox{---}$} \\
\hline
\end{tabular}
\end{table}
\begin{table}[!t]
\caption{Ablation study on TASD-2 reported by mean square error. A lower value $\downarrow$ is better.}
\label{tab:ablation_tasd2}
\centering
\begin{tabular}{|c||c|c|c|}
\hline
 & SyncDiv\_3m & SyncDiv\_10m & Avg.\\
\hline
DTW \cite{berndt1994dtw} & 0.658 & 0.789 & 0.736\\
\hline
2D Correlation \cite{Pedregosa2011scikit-learn}  & 0.763& 0.802& 0.786 \\
\hline
Cross-recurrence \cite{coco2014cross} & 0.682 & 0.747 & 0.721 \\
\hline
AIM \cite{gao2020asymmetric} &  1.535 & 0.515 & 0.919\\
\hline
\hline
TFN & 0.600& 0.520 &0.553 \\
\hline
I3D \cite{carreira2017i3d} &0.373& 0.426& 0.405 \\
\hline
CSM \cite{he2016resnet}  & 0.629 & 0.699 &0.670 \\
\hline
TFN + I3D & 0.384&0.327 &0.350  \\
\hline
TFN + CSM &0.572 &0.506 &0.533 \\
\hline
I3D + CSM & 0.387& 0.373 & 0.378 \\
\hline
TFN+I3D+CSM & \textbf{0.366} &\textbf{0.322} & \textbf{0.340} \\
\hline
\end{tabular}
\end{table}
\subsection{Confusion Matrix}
Table \ref{tab:cm_pt13} reports the row-wisely normalized confusion matrix (CM) for synchrony classification on PT13. 
For consistency, we carefully categorized TASD-2 into three synchrony classes as PT13 with thresholds $\alpha = 7.16 $ and $\beta = 8.36$.
$\alpha$ and $\beta $ were computed from synchrony score distribution in TASD-2  to ensure data splits resemble an approximate normal distribution.
As a result, the percentage of Synchronized, Moderated Synchronized and Unsynchronized data in TASD-2 is $21.3\%, 53.0\% , 25.7\%$.
Table \ref{tab:cm_pt13} reports the CM on TASD-2. Each CM was computed using the best model as reported in Table \ref{tab:ablation_pt13} and \ref{tab:ablation_tasd2}. 
The overall classification accuracy (F1-score) for PT13 and TASD-2 are $91.80\%$ ($0.91$) and $ 78.05\%$ ($0.78$), respectively. 
We achieved a lower accuracy on TASD-2 due to the nature of the dataset and the distribution of scores, which are completely different from PT13. In particular, synchronized and moderately synchronized data are very similar to each other, leading to more errors in these two classes, as shown in Table \ref{tab:cm_tasd2}.


\begin{table}[t]
    \caption{Normalized confusion matrix on PT13 (\%).}
    \label{tab:cm_pt13}
    \centering
    \begin{tabular}{l|l|c|c|c|c|}
    \multicolumn{2}{c}{}&\multicolumn{2}{c}{Prediction}\\
    \cline{3-5}
    \multicolumn{2}{c|}{} & Sync & ModSync & Unsync \\
    \cline{2-5}
    \multirow{2}{*}{Label}& Sync & 97.70 & 1.15 & 1.15 \\
    \cline{2-5}
    &ModSync & 12.36& 85.39& 2.25\\
    \cline{2-5}
   &Unsync & 5.00 & 2.50 & 92.50\\
    \cline{2-5}
    \end{tabular}
\end{table}
\begin{table}[t]
    \caption{Normalized confusion matrix on TASD-2 (\%). }
    \label{tab:cm_tasd2}
    \centering
    \begin{tabular}{l|l|c|c|c|c|}
    \multicolumn{2}{c}{}&\multicolumn{2}{c}{Prediction}\\
    \cline{3-5}
    \multicolumn{2}{c|}{} & Sync & ModSync & Unsync \\
    \cline{2-5}
    \multirow{2}{*}{Label}& Sync &58.62 &41.38 &0 \\
    \cline{2-5}
    &ModSync &3.95 &86.84 & 9.21\\
    \cline{2-5}
   &Unsync &0 &27.78 &72.22\\
    \cline{2-5}
    \end{tabular}
\end{table}

\subsection{Spatial \& Temporal Attention Visualization}
To demonstrate the multi-head self-attention mechanism, we visualized spatial and temporal attention maps of the first layer ($L = 1$) learnt from PT13.
Attention heads for visualization were randomly selected.
The goal of visualizing attention maps in TFN is to show that each head is able to capture distinct motion patterns; such multi-head attention mechanism is not redundant, but rather aids the network in capturing diverse spatial/temporal movement aspects.
It can also illustrate the benefits of TFN in interpreting which specific (spatial) body parts or (temporal) actions are critical in determining movement synchrony when traditional methods like MEA can not. 

\begin{figure}[t]
    \centering
    \includegraphics[width = 0.75\columnwidth]{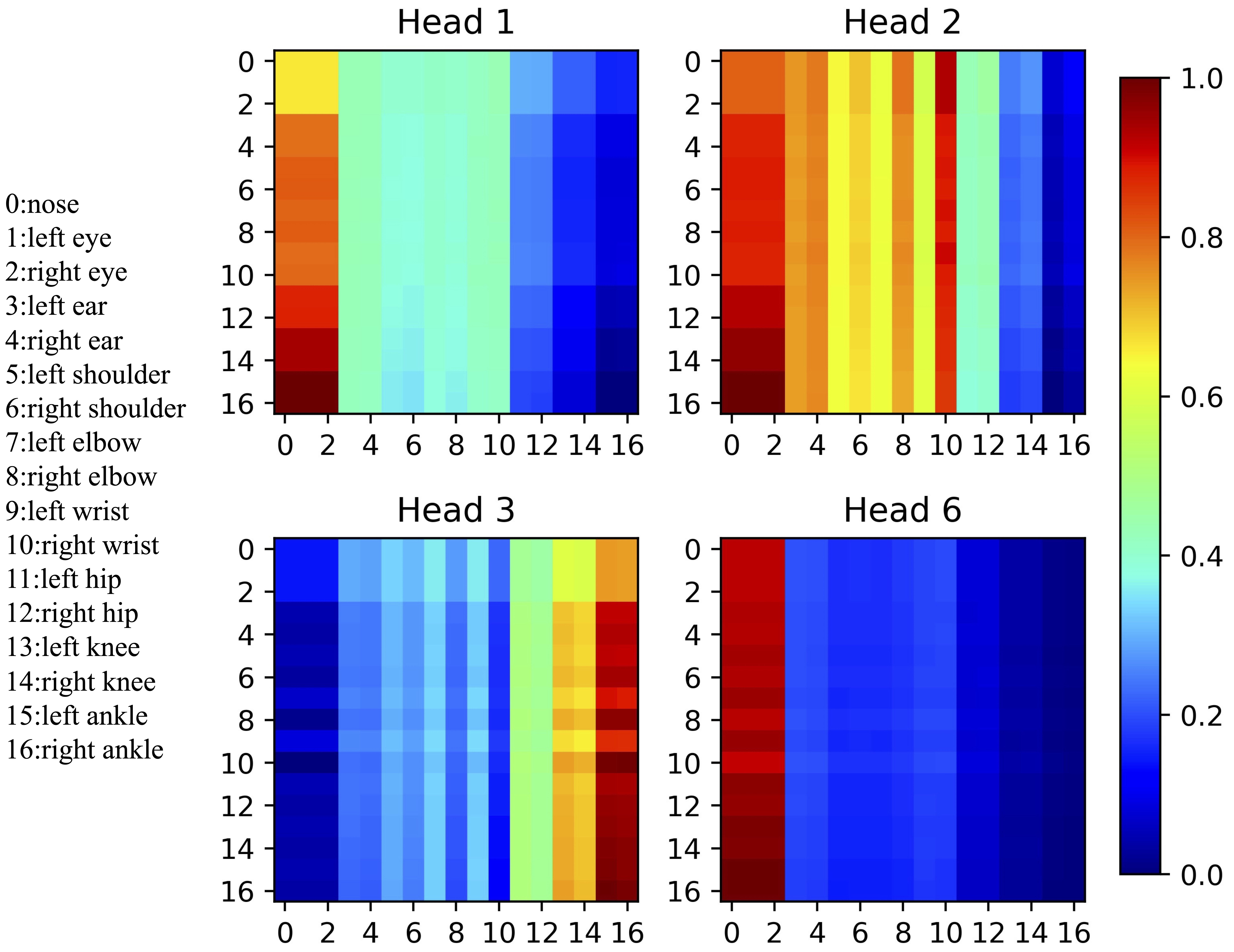}
    \caption{Visualization of self-attentions in the spatial transformer from PT13. The pixel $w_{i,j}$ ($i$: row, $j$: column) denotes the attention weight $w$ of the $j$-th query (joint $j$) for the $i$-th key (joint $i$). Red color indicates larger attention weight. The attention map has been normalized to a range of 0 to 1.}
    \label{fig:spatial_attn}
\end{figure}

\textbf{Spatial attention} computed from PT13 is visualized in Fig. \ref{fig:spatial_attn}, where the $x (y)$-axis corresponds to the query (key) of $J=17$ joints, and the pixel value at $(x,y)$ indicates the attention density. 
We noticed that each attention head returns distinct attention intensities, indicating the heterogeneity in local connections among disparate groups of joints learned by the network.
For instance, Head 1 and Head 6 are concerned with upper body parts (joint 0 $\sim$ 2), whereas Head 3 examines lower body connections (joint 13 $\sim$ 16), and Head 2 focuses on the upper and middle body (joint 0 $\sim$ 4, 6, 8, 10) interactions. 



\begin{figure}[!h]
    \centering
    \includegraphics[width = 0.68\columnwidth]{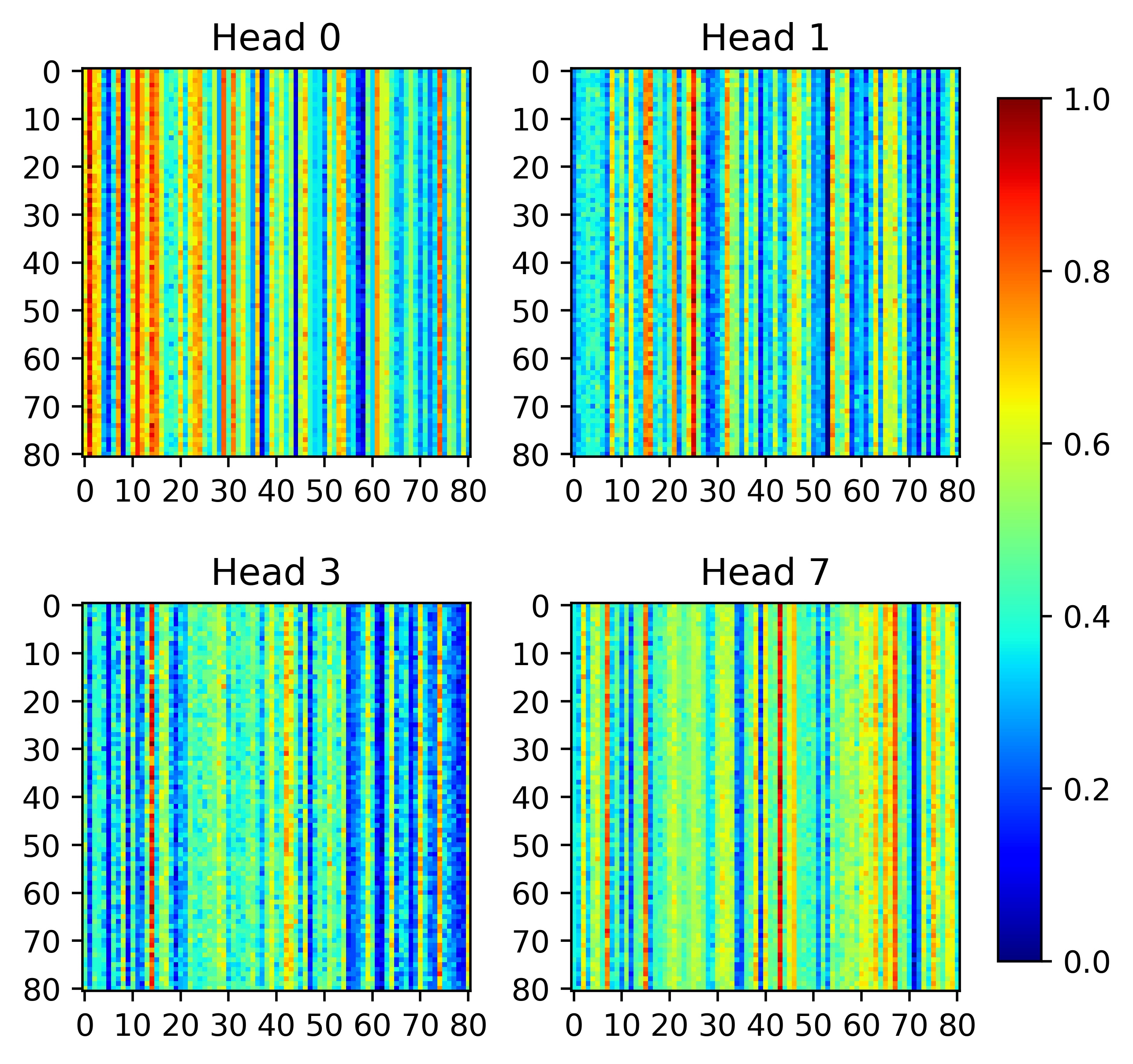}
    \caption{Visualization of self-attentions in the temporal transformer from PT13. The pixel $w_{i,j}$ ($i$: row, $j$: column) denotes the attention weight $w$ of the $j$-th query (frame $j$) for the $i$-th key (frame $i$). Red color indicates larger attention weight. The attention map has been normalized to a range of 0 to 1.}
    \label{fig:temp_attn}
\end{figure}

\textbf{Temporal attention} computed from PT13 is visualized in Fig. \ref{fig:temp_attn},
where the $x (y)$-axis corresponds to the query (key) of $f = 81$ frames, and the pixel value at $(x,y)$ indicates the attention density.
We observed that different attention heads have learnt diverse long-term global dependencies.
For example, Head 0 mainly focuses on early frames (between 0 and 25), while Head 7 catches the relationship between frames 7, 15, 42, 46, and 67 despite significant temporal disparities. 

\section{Discussion}
\label{sec:discussion}
Based on the input, the three sub-networks of our framework can be divided into two classes: (1) skeleton-based approach (TFN \& CSM) and (2) optical-flow-based approach (I3D).
Under privacy-preserving conditions, each class has unique pros and cons.
 
In both TASD-2 and PT13 datasets, the skeleton-based approach exhibits good generality and interpretability.
TFN, for instance, demonstrates a strong ability to capture both local and global joint correlations, producing good results across both datasets.
However, skeleton-based approaches are susceptible to data noise caused by pose detector failures, especially when challenging atypical poses are encountered.
Furthermore, certain critical scenes can be ignored by skeleton-based methods when a person is not present, such as the splash at the end of a dive, which is also critical for synchrony assessment.

The optical-flow-based approach manages to retain original motion features from raw videos and achieves the best results among all three sub-networks on TASD-2. 
Nevertheless, it is susceptible to background noise caused by the movement of other objects in the same scene. 
For example, many scenes in PT13 are animated by an actively moving robot serving as a therapeutic tool.
As a consequence, this method does not perform well on PT13.
Furthermore, 3D convolutional networks require more computational resources and take longer to train due to network architecture and input data volume. 
For example, optical flow data derived from frames is approximately 1,200 times larger than skeleton data in the same scene, since the latter consists merely of joint coordinates.




\section{Conclusion}
\label{sec:conclusion}
Movement synchrony estimation has been extensively applied in multiple fields  including sports, physical therapy, and rehabilitation.
However, privacy concerns were not adequately addressed.
This paper proposed an ensemble network for movement synchrony assessment under privacy-preserving conditions. 
Our framework is entirely based on secondary data that is both identity-agnostic and privacy-preserving, such as skeletal data and optical flow.
Extensive experiments on TASD-2 and PT13 datasets demonstrate the effectiveness of the proposed framework.

In future work, we will integrate graph neural networks to generate spatial skeletal representations \cite{plizzari2021skeletonSTTF}. 
We will also investigate other privacy-preserving data modalities, such as gaze data \cite{guo2021icmi}, facial landmarks, and emotion clues \cite{li21twostage}.







\bibliographystyle{IEEEtran}
\bibliography{ref}
\end{document}